\documentclass[10pt,twocolumn,letterpaper]{article}

\usepackage[pagenumbers]{cvpr} 

\definecolor{cvprblue}{rgb}{0.21,0.49,0.74}
\usepackage[pagebackref,breaklinks,colorlinks,allcolors=cvprblue]{hyperref}

\usepackage{url}
\usepackage{graphicx} 
\usepackage{booktabs}
\usepackage{times}
\usepackage{epsfig}
\usepackage{graphicx}
\usepackage{amsmath}
\usepackage{amssymb}
\usepackage{pifont}
\usepackage{booktabs}
\usepackage{arydshln}
\usepackage{color}
\usepackage{colortbl}
\usepackage{subcaption}
\usepackage{multirow}
\usepackage{float}
\usepackage{rotfloat}
\usepackage{capt-of}
\usepackage{longtable}
\usepackage{diagbox}
\usepackage{makecell}
\usepackage{pgfplots}
\usepackage{xspace}
\usepackage{wrapfig}
\usepackage{enumitem}
\usepackage{fontawesome}
\usepackage{epigraph}
\usepackage{multirow}
\usepackage{multicol}
\usepackage{rotating}
\usepackage{amssymb}
\usepackage{pifont}
\usepackage{algpseudocode}
\definecolor{mycolor_blue}{HTML}{E7EFFA}
\definecolor{mycolor_green}{HTML}{E6F8E0}
\definecolor{mycolor_gray}{HTML}{ECECEC}
\definecolor{pearDark}{HTML}{2980B9}

\usepackage{fbox}
\definecolor{textcolor1}{rgb}{0.25,0.5,0.5}
\definecolor{textcolor2}{rgb}{0.7,0.25,0.25}
\definecolor{linkc}{rgb}{0, 0.44, 0.74}
\definecolor{eqc}{rgb}{1, 0, 0}
\definecolor{myy}{RGB}{126,95,0}
\definecolor{mygray}{gray}{.9}
\definecolor{bblue}{RGB}{30,80,120}
\definecolor{mygray1}{gray}{.7}
\definecolor{ggray}{RGB}{127,127,127}
\definecolor{mygreen}{RGB}{93,174,86}
\definecolor{citecolor}{HTML}{229954}

\usepackage{xcolor}

\usepackage[capitalize]{cleveref}
\crefname{section}{Section}{Secs.}
\Crefname{section}{Section}{Sections}
\Crefname{table}{Table}{Tables}
\crefname{table}{Tab.}{Tabs.}

\usepackage{indentfirst}
\usepackage{utfsym}

\def\paperID{****} 
\def\confName{CVPR}
\def\confYear{2025}

\title{ASGDiffusion: Parallel High-Resolution Generation with \\Asynchronous Structure Guidance}

\author{Yuming Li$^{1}$, Peidong Jia$^{1}$, Daiwei Hong$^{1}$, Yueru Jia$^{1}$, Qi She$^{2}$, Rui Zhao$^{3}$, Ming Lu$^{4}$, Shanghang Zhang$^{1,*}$ \\
    $^1$Peking University \\
    $^2$ByteDance Inc. \\
    $^3$Tencent Robotics X, Shenzhen, China \\
    $^4$Intel Labs China}

\begin{document}

\twocolumn[{
\renewcommand\twocolumn[1][]{#1}
\maketitle
\begin{center}
    \captionsetup{type=figure}
    \includegraphics[width=\textwidth]{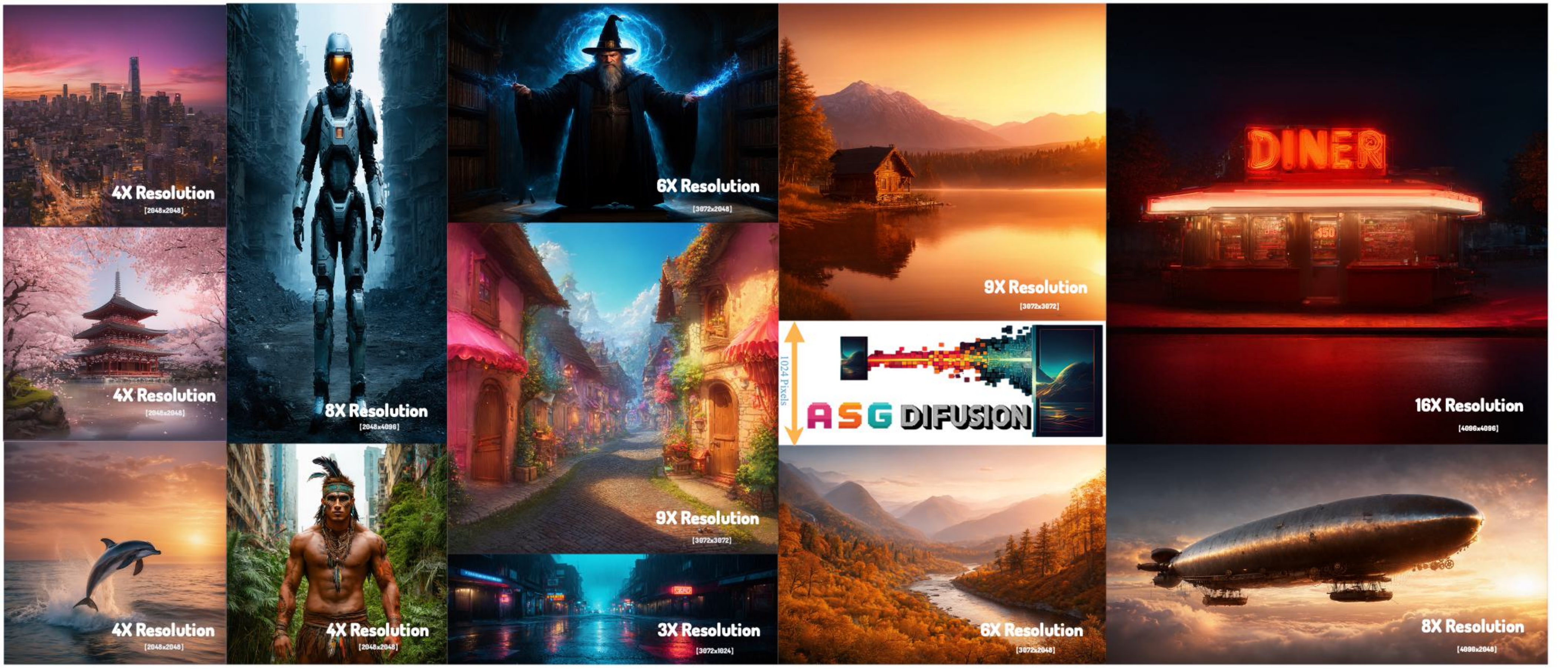}
    \captionof{figure}{The generated samples of ASGDiffusion based on Stable Diffusion 3 (SD3). While SD3 can synthesize images up to 1024x1024, our method enhances SD3's capability to generate images at resolutions exceeding 1024x1024 without requiring fine-tuning or high memory usage. Best viewed by zooming in.}
\end{center}
}]
\begingroup\renewcommand\thefootnote{\fnsymbol{footnote}}
\footnotetext[1]{* Corresponding author.}
\endgroup
\begin{abstract}
Training-free high-resolution (HR) image generation has garnered significant attention due to the high costs of training large diffusion models. Most existing methods begin by reconstructing the overall structure and then proceed to refine the local details. Despite their advancements, they still face issues with repetitive patterns in HR image generation. Besides, HR generation with diffusion models incurs significant computational costs. Thus, parallel generation is essential for interactive applications. To solve the above limitations, we introduce a novel method named ASGDiffusion for parallel HR generation with Asynchronous Structure Guidance (ASG) using pre-trained diffusion models. To solve the pattern repetition problem of HR image generation, ASGDiffusion leverages the low-resolution (LR) noise weighted by the attention mask as the structure guidance for the denoising step to ensure semantic consistency. The proposed structure guidance can significantly alleviate the pattern repetition problem. To enable parallel generation, we further propose a parallelism strategy, which calculates the patch noises and structure guidance asynchronously. By leveraging multi-GPU parallel acceleration, we significantly accelerate generation speed and reduce memory usage per GPU. Extensive experiments demonstrate that our method effectively and efficiently addresses common issues like pattern repetition and achieves state-of-the-art HR generation. 
\end{abstract}

\section{Introduction}
\label{sec:intro}
Diffusion models demonstrate remarkable capabilities in generating high-quality images, making them a favored option across various applications. Despite these capabilities, training a high-resolution diffusion model requires significant computational resources. For instance, it has been reported that training Stable Diffusion 3 takes over 24 days using 256 A100 GPUs \cite{emostaque2024}. This process requires substantial GPU power and access to large datasets, making it both time-consuming and costly. Additionally, this is solely for training at a resolution of 1024x1024; the resources needed for training at higher resolutions would increase exponentially and be nearly limitless. Therefore, training-free HR image generation has gained significant interest.

Recent advances in training-free high-resolution diffusion methods, such as MultiDiffusion \cite{bar2023multidiffusion}, ScaleCrafter \cite{he2023scalecrafter}, DemoFusion \cite{du2024demofusion}, and CutDiffusion \cite{lin2024cutdiffusion} have made significant progress. MultiDiffusion employs overlapping high-resolution patches but struggles with maintaining global consistency and preventing repetitive objects. ScaleCrafter generates full images using dilated convolutions to maintain global consistency; however, this method restricts generative capacity, resulting in structural distortions and repetitive patterns. DemoFusion, by incorporating Progressive Upscaling, Skip Residual, and Dilated Sampling mechanisms, generates higher quality images but at the cost of requiring more inference steps, significantly increasing the generation time. CutDiffusion shuffles latent noises to generate high-resolution images; however, it fails to address pattern repetition and does not support multi-GPU parallel processing, which could accelerate generation speed. In summary, current methods are largely impeded by repeated patterns, which significantly deteriorate the overall image quality. These methods lack support for multi-GPU parallel acceleration, limiting their efficiency and scalability.

To address these challenges, we introduce ASGDiffusion, an innovative parallel method for generating high-resolution images. ASGDiffusion is a two-stage, patch-based approach that first constructs a consistent global structure and then refines the details to create a high-quality image. In the first stage, we use the first patch that acts as global structure guidance, ensuring that all patches maintain consistent global structure throughout the generation process. Besides, after analyzing the cross-attention heatmaps, we found that object regions attract more attention than background regions. To leverage this, we utilize cross-attention heatmaps to create a weight mask that adjusts structure guidance. This minimizes background interference while preserving overall consistency in object areas.

However, waiting for structure guidance at each time step introduces communication overhead. To enable parallel generation, we further propose a parallelism strategy to calculate the patch noises and structure guidance asynchronously. Instead of synchronously waiting for the structure guidance at each time step, we use guidance from the previous time step ($t-1$) for the current denoising step ($t$). Due to minimal changes between consecutive steps, this asynchronous approach allows for overlapping communication and computation, thereby reducing latency and enhancing parallel efficiency. 

ASGDiffusion can be easily integrated into various versions of Stable Diffusion, including SD1.5, SD2.1, SDXL, and SD3, significantly enhancing the quality and efficiency of high-resolution image generation. Our method ensures consistently high-resolution images while significantly reducing generation time compared to other approaches. The main contributions are summarized as follows:
\begin{itemize}
\item We present ASGDiffusion, an innovative training-free method for high-resolution image generation that addresses pattern repetition through  structure guidance, which is weighted by an attention mask.

\item We develop a strategic multi-device parallel acceleration method to calculate patch noises and structure guidance asynchronously, which significantly speeds up generation and reduces memory usage for each device.

\item We utilize ASGDiffusion across various versions of Stable Diffusion, highlighting the benefits of our approach compared to existing methods.
\end{itemize}

\begin{figure}
    \centering
    \includegraphics[width=1\linewidth]{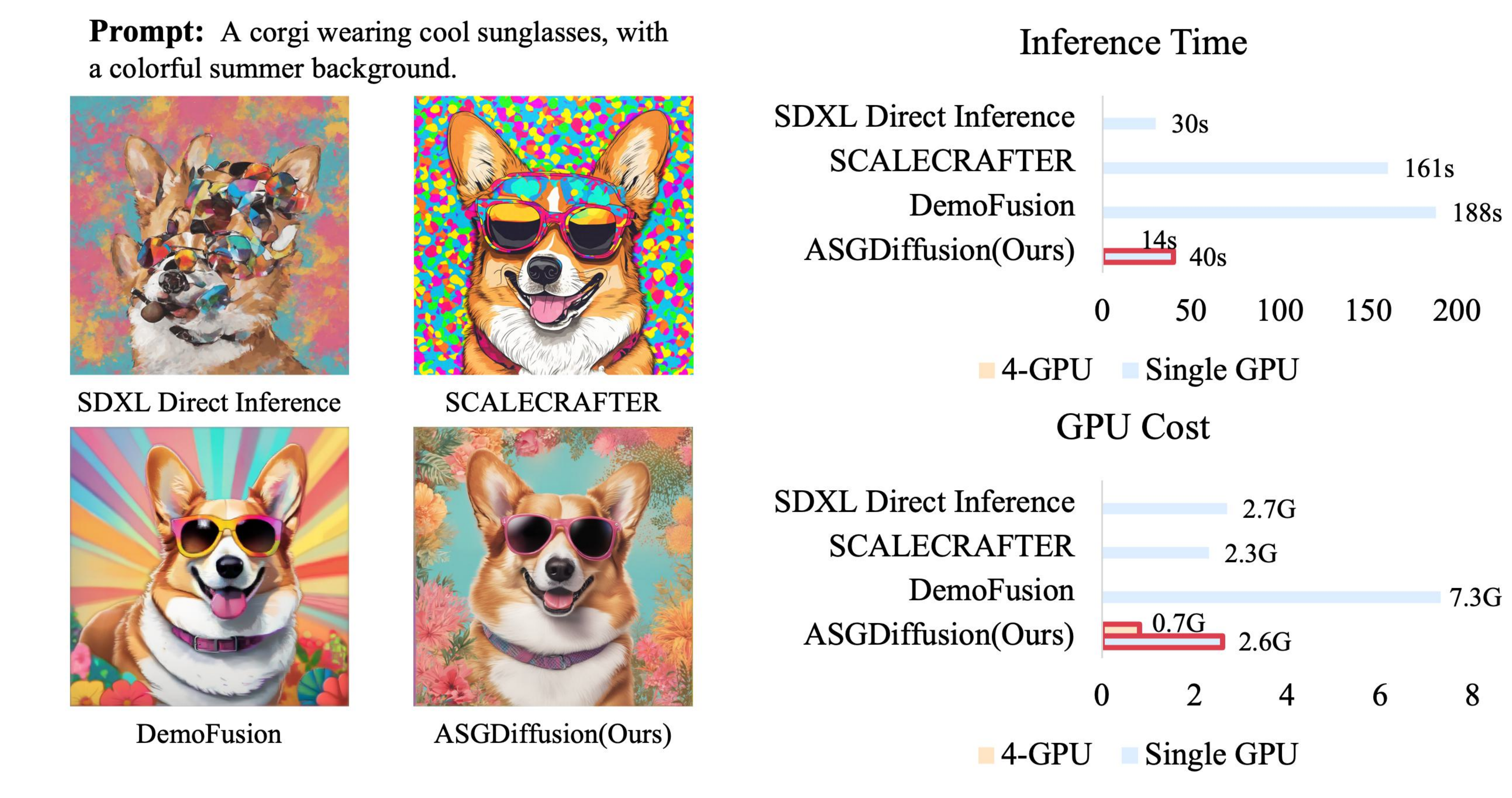}
    \caption{The comparison of generated images, inference time, and GPU cost for different methods at 2048x2048 resolution on RTX 4090. Our method (ASGDiffusion) is the fastest and supports parallel processing.}
    \vspace{-0.35cm}
    \label{fig3}
\end{figure}

\section{Related work}

\subsection{Single image super-resolution (SISR)}
Deep learning methods have revolutionized Single Image Super-Resolution (SISR). Early neural network-based approaches like SRCNN \cite{dong2015image}, VDSR \cite{kim2016accurate}, and ESPCN \cite{shi2016real} demonstrated significant performance improvements. More advanced networks, such as SRGAN \cite{ledig2017photo}, EDSR \cite{lim2017enhanced}, and BSRGAN \cite{zhang2021designing}, further enhanced both image quality and efficiency. Shi et al. \cite{shi2016real} introduced the sub-pixel convolution layer, which effectively rearranges pixels from low-resolution inputs to generate high-resolution outputs.

\subsection{Diffusion models}
Diffusion models add noise to data in a forward process and then learn to reverse this process to generate samples. Examples include Denoising Diffusion Probabilistic Models (DDPM) \cite{ho2020denoising} and Denoising Diffusion Implicit Models (DDIM) \cite{song2020denoising}, which have shown success in various tasks. Latent Diffusion Models (LDMs) operate in latent spaces, leading to efficient high-quality image generation \cite{rombach2022high}.

\subsection{Training-free high-resolution generation}
Previous studies fall into two categories: training-based methods, such as Cascaded Diffusion Models \cite{ho2022cascaded} and Relay Diffusion \cite{teng2023relay}, and training-free methods, including ScaleCrafter \cite{he2023scalecrafter} MultiDiffusion \cite{bar2023multidiffusion} and DemoFusion \cite{du2024demofusion}. ScaleCrafter \cite{he2023scalecrafter} utilizes dilated convolutions to expand the receptive field, effectively reducing object repetitiveness but potentially introducing structural distortion and degrading local detail at higher resolutions. Patch-based methods like MultiDiffusion \cite{bar2023multidiffusion} split high-resolution images into smaller patches for processing, combining multiple diffusion paths to maintain consistency. DemoFusion \cite{du2024demofusion} enhances generation by incorporating global semantic information and using skip residual connections and dilated sampling. However, it still encounters challenges with repetitive objects and chaotic local details, alongside a longer generation time due to increased inference steps. Recently, DiffuseHigh \cite{kim2024diffusehigh} generates high-resolution images by progressively refining low-resolution inputs, but depends on input quality,Upsample Guidance \cite{hwang2024upsample} has been introduced to adapt pre-trained diffusion models for higher resolutions with minimal adjustments, eliminating the need for additional training.

\section{Background}
\subsection{Diffusion models}
Diffusion models transform an original sample \(x_0\) into progressively noised versions \(x_t\) until reaching pure noise \(x_T\). Most models follow the framework of Denoising Diffusion Probabilistic Models (DDPMs)~\cite{ho2020ddpm}, using Gaussian noise:
\begin{equation}
    x_t = \sqrt{\alpha_t} x_0 + \sqrt{1 - \alpha_t} \epsilon_t,
\end{equation}
where \(\epsilon_t \sim \mathcal{N}(0, I)\) and \(\alpha_t\) is a noise schedule decreasing over time. The generation process is achieved through a backward diffusion method that utilizes a noise predictor \(\epsilon(x_t, t)\). This predictor typically employs a U-Net architecture~\cite{ronneberger2015unet} for greater adaptability across various resolutions.

\subsection{Guidance techniques for diffusion models}
Conditional sampling techniques have been created to guide the generation process. Recent study \cite{ho2022classifier} introduced a guidance method that integrates the gradient of the log probability from a classifier into \(\epsilon(x_t, t)\), enabling class-conditioned image generation.

\textbf{Classifier-Free Guidance (CFG)} was subsequently proposed to eliminate the need for an external classifier by modifying the noise predictor to directly accept a condition \(c\). The predicted noise under CFG is given by:
\begin{equation}
    \tilde{\epsilon}(x_t, t; c) = \epsilon(x_t, t) + w \left[ \epsilon(x_t, t; c) - \epsilon(x_t, t) \right],
\end{equation}
where \(w\) is the guidance scale. Proper adjustment of \(w\) can significantly enhance the fidelity and alignment of the generated images, ensuring they adhere to given prompts while maintaining image quality.

\begin{figure*}[ht]
    \centering
    \includegraphics[width=\textwidth]{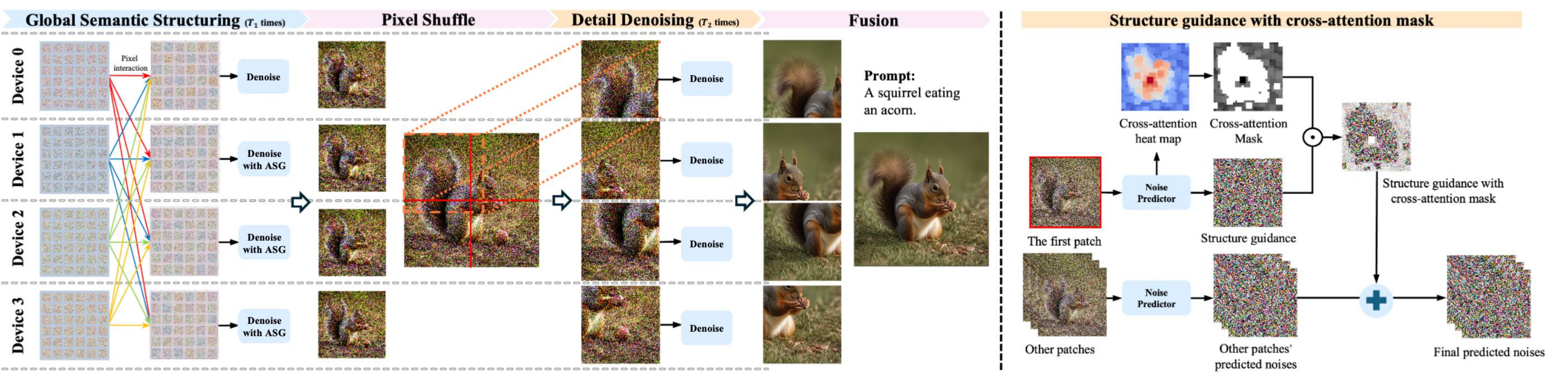}
    \caption{The pipeline of ASGDiffusion. Following recent works, our method also consists of two stages. In the first stage, we refine the overall structure with the proposed asynchronous structure guidance(ASG). In the second stage, we recover the details to produce the final image. Right is the illustration of structure guidance with the cross-attention mask. We introduce a parallelism strategy to make the structure guidance asynchronous, allowing multi-GPU parallel acceleration.}
    \label{fig:pipeline}
\end{figure*}

\section{Method}
\subsection{Overview}
Recent studies \cite{lin2024cutdiffusion} indicate that diffusion models prioritize constructing the semantic structure during the initial phases of denoising while they focus on refining fine details in the later stages. Following recent works \cite{lin2024cutdiffusion}, our method is divided into two stages. 

In the first stage, we aim to construct a consistent overall structure. CutDiffusion employs a pixel interaction operation where pixels in the same positions across different patches are randomly exchanged to maintain the overall structure. Although pixel interaction enables patches to share information, which reduces the issue of pattern repetition while preserving the Gaussian distribution of each patch, we find the pixel interaction may still contain obvious pattern repetition. To address this, we introduce \textbf{structure guidance with cross-attention mask} to further refine the semantic structure in Sec.~\ref{method_sg}. In the second stage, we also refine the details to produce the final image.

For both stages, directly denoising the HR latent noises would be computationally expensive. Dividing the HR latent noises into multiple LR patch noises enables parallel generation. However, the LR patch noise must await structure guidance before denoising at each timestep, which limits parallel capacity. Therefore, we propose \textbf{asynchronous structure guidance} that enables each patch to perform denoising independently without waiting for the most recent structure guidance in Sec.~\ref{method_asg}. Finally, the denoised patches are fused to form the final high-resolution image. This parallelism strategy effectively reduces computational overhead per GPU while maintaining consistency and image quality. The complete pipeline is illustrated in Fig.~\ref{fig:pipeline}.

\subsection{Structure guidance with cross-attention mask}\label{method_sg}
As mentioned above, the pixel interaction of CutDiffusion \cite{lin2024cutdiffusion} still suffers from pattern repetition problems, as shown in Fig. \ref{fig6}. We hypothesized that this issue resulted from insufficient global semantic guidance during the initial stage of denoising. To address this issue, we introduced a structured guidance to enhance the consistency of semantic structures throughout the entire image.

As shown in Fig.~\ref{fig:pipeline}, the structural guidance is created by selecting the first low-resolution patch noise to represent the overall structure of the high-resolution image. In each denoising step, we combine the predicted noise from the first patch with noise predictions from other patches to maintain the overall semantic structure. Specifically, the final predicted noise $\tilde{\epsilon}(x_t^{(i)}, t)$ for other patches can be formulated as:
\begin{equation}
\tilde{\epsilon}(x_t^{(i)}, t) = \epsilon(x_t^{(i)}, t) + w_t [\epsilon(x_t^{\text{0}}, t) - \epsilon(x_t^{(i)}, t)],
\end{equation}
where $\epsilon(x_t^{(i)}, t)$ represents the original noise prediction for patch $i$, and $\epsilon(x_t^{\text{0}}, t)$ is the noise predicted of the first patch, which will be used as structure guidance. The parameter $w_t$ regulates the influence of structure guidance, ensuring that other patch noises are modified to align with the global structure provided by the first patch.

After incorporating structure guidance, as shown in Fig.~\ref{fig6}, we observed a significant enhancement in the object regions within the image, effectively eliminating the pattern repetition issue. However, a new issue has arisen: the background areas are showing signs of blurriness and deterioration. We hypothesized that this discrepancy might be attributed to a mismatch in attention: while the object regions received significant attention from the cross-attention mechanism, background regions were relatively overlooked. Furthermore, as previously mentioned, diffusion models often enhance details in the later stages. This means that guidance in the early stages may disrupt probability distributions in low-attention areas, such as the background. 

To further investigate, we visualize the cross-attention maps during different stages of denoising. As shown in Fig.~\ref{cross-attention}, the cross-attention maps indicated that during the early stages of denoising, the cross-attention was dispersed and lacked sharp focus, as the semantic structure of the image had not yet fully developed. As denoising progressed, we observed that attention areas concentrated on the main objects in the scene, such as the squirrel, indicating a stronger and more consistent semantic focus. We also visualized the cross-attention across different layers of the U-Net model to gain deeper insight, specifically comparing the downsampling layers and the upsampling layers. The results indicated that the upsampling layers of the U-Net demonstrated more focused cross-attention compared to the downsampling layers. This is because the upsampling layers refine image details and align semantic information during reconstruction, resulting in greater attention to key objects. In contrast, downsampling layers mainly focus on feature extraction and capturing broader contextual information, which results in a more distributed attention distribution. Thus, the cross-attention in upsampling layers is more effective as a mask, as it highlights significant objects while preserving semantic coherence.

\begin{figure}[t]
    \includegraphics[width=1\linewidth]{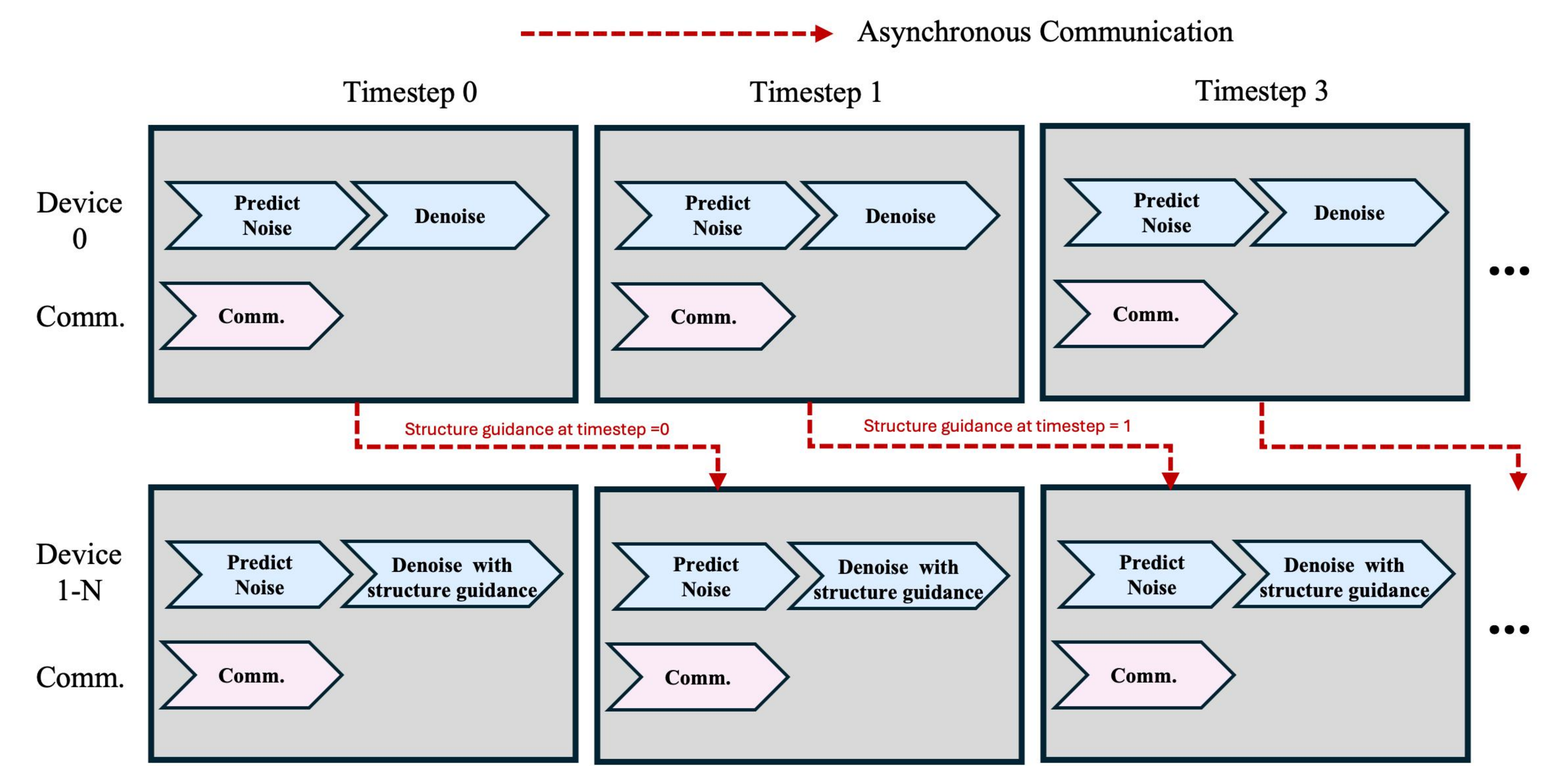}
    \vspace{-0.2cm}
    \caption{Timeline visualization of asynchronous structure guidance(ASG). Comm. means communication. The Comm. overhead is fully hidden within the computation.} 
    \label{ASG}
\end{figure}

Considering these findings, we propose using the cross-attention heat map as a mask to filter the structural guidance, especially in areas with low attention. Specifically, we normalize the attention heat map and use it to modulate the guidance as a weight.
By doing this, we can selectively apply guidance from the structure to regions with high attention, minimizing the impact on low-attention areas like the background. The cross-attention mask effectively retains background details while ensuring global consistency throughout the image. The final structure guidance, which includes a cross-attention mask, is formulated as follows:
\begin{equation}
\tilde{\epsilon}(x_t^{i}, t) = \epsilon(x_t^{i}, t) + w_t M[\epsilon(x_t^{\text{0}}, t) - \epsilon(x_t^{i}, t)],
\end{equation}
where the cross-attention mask $M$ adjusts how the structure affects the noise from other patches. By integrating structural guidance with a cross-attention mask, we achieve both global consistency and clear background details.

\subsection{Asynchronous structure guidance}\label{method_asg}
Generating high-resolution (HR) images with diffusion models requires significant computational resources, making efficient parallel generation essential for interactive applications. In our approach, we use structure guidance with a cross-attention mask to ensure consistency across LR patch noises. Therefore, the LR patch noise must await structure guidance before denoising at each timestep, which limits the parallel capacity.

To address this, we propose an asynchronous structure guidance that integrates synchronization with computation, allowing for parallel acceleration without delays. Rather than relying on structural guidance from the current time step ($t$), our method utilizes structural guidance from the previous time step ($t-1$) to directly generate the denoised patches for the current time step ($t$). This asynchronous approach leverages the similarity found in consecutive time steps of diffusion models, enabling devices to reuse slightly outdated guidance while ensuring semantic coherence. By utilizing guidance from step $t-1$, the current step can start denoising immediately, eliminating synchronization delays and greatly enhancing parallel efficiency. The complete procedure is outlined in Fig.~\ref{ASG}, where $G_t$ denotes the guidance at time step $t$.:
\begin{equation}
G_t = w_t M \left[\epsilon(x_{t-1}^{\text{0}}, t - 1) - \epsilon(x_t^{(i)}, t)\right]
\end{equation}
\begin{equation}
\tilde{\epsilon}(x_t^{(i)}, t) = \epsilon(x_t^{(i)}, t) + G_t
\end{equation}

Our experiments demonstrate that the proposed asynchronous structure guidance is effective. By utilizing our method, we only need to denoise individual patches throughout the entire process instead of the whole high-resolution latent. This allows for efficient parallel generation across multiple devices, something that traditional high-resolution methods often struggle to accomplish. Compared to synchronous approaches, our method reduces communication overhead, increases generation speed, and maintains high-quality image outputs compared to synchronous approaches.

\begin{figure*}[ht]
    \centering
    \includegraphics[width=1\textwidth]{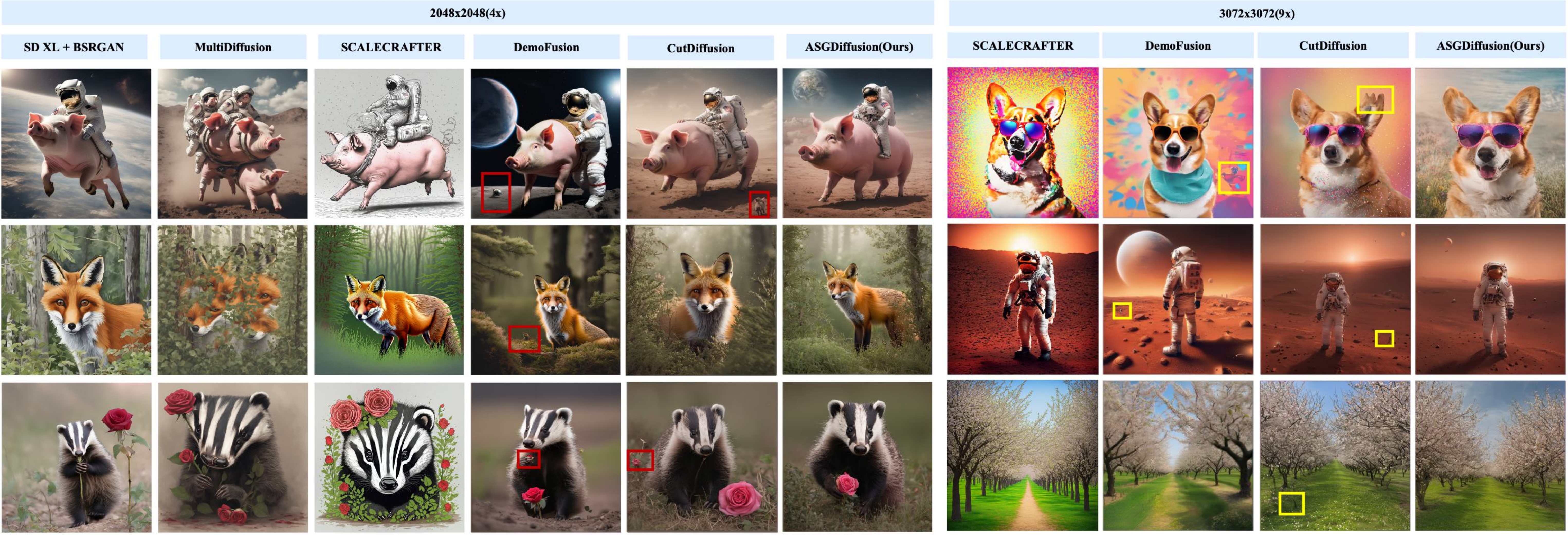}
    \caption{Comparison of different methods. (a) SDXL+BSRGAN, (b) MultiDiffusion, (c) ScaleCrafter, (d) DemoFusion, (e) CutDiffusion, (f) ASGDiffusion (Ours). MultiDiffusion, ScaleCrafter, and DemoFusion fail to solve the pattern repetition problem in HR generation. Our method, ASGDiffusion, refines the overall structure by the structure guidance. Additionally, we propose a parallelism strategy to make the structural guidance asynchronous, enabling multi-GPU acceleration.}
    \label{fig:comparison}
\end{figure*}

\section{Experiments}

\subsection{Experimental setup}
We conducted the evaluation experiments on the text-to-image model, Stable Diffusion (SD) XL 1.0 \cite{podell2023sdxl}, generating multiple higher resolutions beyond the training resolution. Our method can also be easily integrated into other versions of Stable Diffusion, including SD 1.5, SD 2.1, and SD 3. We compared our method with several representative generative approaches: SDXL Direct Inference, SDXL+BSRGAN, ScaleCrafter \cite{he2023scalecrafter}, MultiDiffusion \cite{bar2023multidiffusion}, CutDiffusion \cite{lin2024cutdiffusion} and DemoFusion \cite{du2023demofusion}. The experiments were conducted on NVIDIA RTX 4090 GPU. For all methods, we used a denoising schedule consisting of 50 steps, with both the first and second stages requiring 25 steps each.

\subsection{Inference time}
Tab.~\ref{tab:inference_time} demonstrates the significant state-of-the-art advantage of our method in terms of generation speed.
ScaleCrafter experiences considerable time overhead due to its use of dilated convolutions and direct denoising of high-resolution noise. For patch-wise inference approaches, MultiDiffusion requires more time because it needs to denoise a larger number of patches. Demofusion, with its progressive upscaling strategy, increases inference time due to the additional steps required for denoising. CutDiffusion is faster than our method without multi-GPU parallelism because it does not require the computation of cross-attention mask. However, CutDiffusion still suffers from the pattern repetition problem, as shown in Fig.~\ref{fig:comparison}.

In contrast, ASGDiffusion demonstrates time efficiency and high-quality generation. When using 4 GPUs, ASGDiffusion operates 13.4 times faster than Demofusion at a resolution of 2048 × 2048 (14 seconds compared to 188 seconds). Furthermore, our method achieves a 2.4× speedup on 4 GPUs compared to a single GPU, processing the same resolution in 14 seconds instead of 34 seconds. These results demonstrate the remarkable efficiency of ASGDiffusion, especially in tasks that require rapid high-resolution image generation, making it a highly effective solution for practical applications.

\begin{figure}[t]
    \includegraphics[width=1\linewidth]{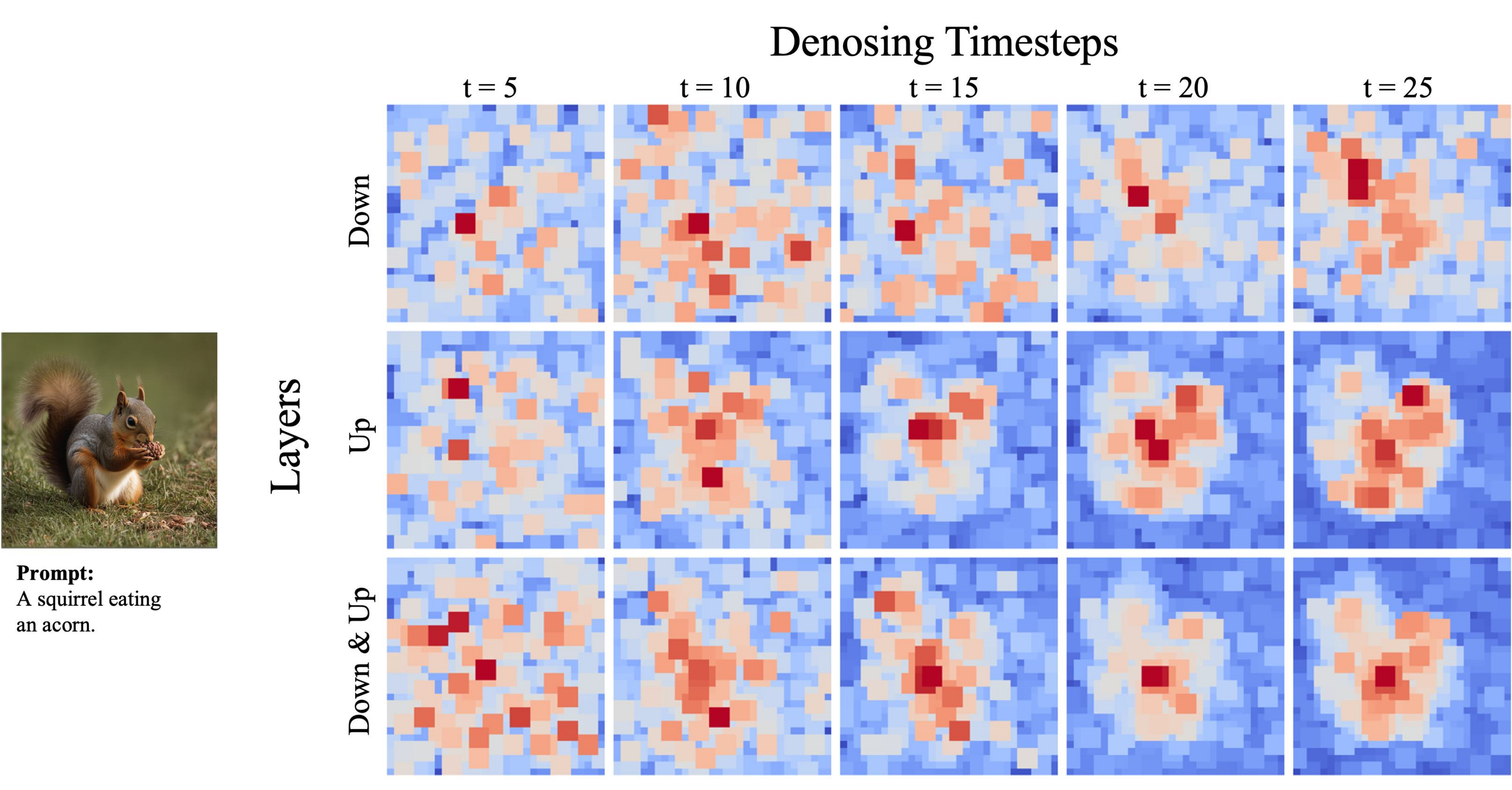}
    \caption{Cross-attention heatmap visualization.} 
    \label{cross-attention}
\end{figure}

\begin{table}[t]
    \centering
    \tiny
    \setlength{\tabcolsep}{2pt} 
    \caption{The inference time of recent training-free HR generation methods and ASGDiffusion across various resolutions. Importantly, our method is unique in its support for multi-GPU parallelism, which is made possible by the proposed asynchronous structure guidance.}
    \label{tab:inference_time}
    \begin{tabular}{lcccc}
    \hline
    Method & 1024 $\times$ 2048 & 2048 $\times$ 2048 & 3072 $\times$ 3072 & 4096 $\times$ 4096 \\ \hline
    SDXL  & 14s & 30s & 95s & 240s \\
    MultiDiffusion & 34s & 110s & 275s & 840s \\
    ScaleCrafter & 25s & 161s & 260s & 584s \\
    DemoFusion & 40s & 188s & 666s & 1380s \\
    CutDiffusion  & 13s & 32s & 114s & 258s \\ 
    ASGDiffusion (ours)  \textbf{1GPU} & \textbf{18s} & \textbf{40s} & \textbf{132s} & \textbf{294s} \\

    ASGDiffusion (ours) \textcolor{red}{4GPU} & \textcolor{red}{10s} & \textcolor{red}{14s} & \textcolor{red}{59.4s} & \textcolor{red}{105s} \\ \hline
    \end{tabular}
\end{table}

\subsection{Qualitative evaluation}
Fig.~\ref{fig:comparison} presents a visual comparison of our method with other approaches, each producing images at a resolution of 2048 × 2048 and 3072 × 3072. SDXL+BSGAN effectively preserves the correct semantic structure, but this super-resolution model simply produces high-resolution images that closely resemble the content of the low-resolution input. Consequently, the generated images often show excessive smoothing and lack essential details needed to achieve the desired high-resolution visual effects. MultiDiffusion lacks global semantic structure guidance, resulting in repetitive content generation within the images. ScaleCrafter offers a solution to the pattern repetition problem found in MultiDiffusion. Nonetheless, the use of dilated convolution kernels affects the quality of the produced images. As the results show, it not only changes the style of the generated images but also causes certain elements, like roses, to be generated repetitively. Demofusion creates images with enhanced local details, such as the fur of badgers and corgis, as well as foliage and various natural elements. The semantic structure of the generated images is quite robust. However, it tends to produce some minor repetitions of objects, such as extra foxes, roses, astronauts, and corgi heads in the images. CutDiffusion utilizes pixel interaction to maintain the overall structure. However, this interaction still exhibits noticeable pattern repetition.

In contrast to the compared methods, ASGDiffusion generates images with a correct and globally consistent semantic structure, eliminating any repetition of minor objects. It also excels at depicting local details, such as the fur of animals and the flowers on trees. Overall, ASGDiffusion excels in maintaining global semantic consistency while also ensuring high quality in local detail.

\begin{table}[t]
    \centering
    \tiny  
    \caption{Quantitative comparison results. The best results are marked in \textbf{bold}, and the second best results are marked by \underline{underline}.}
    \label{tab:quantitative_comparison}
    \begin{tabular}{lcccccc}
    \hline
    Resolution & Method & FID $\downarrow$ & IS $\uparrow$ & FID$_c$ $\downarrow$ & IS$_c$ $\uparrow$ & CLIP $\uparrow$ \\ \hline

    \multirow{6}{*}{1024$\times$2048} 
    & SDXL + BSRGAN   & \underline{64.39} & \underline{13.75} & \textbf{41.32} & \underline{19.64} & 30.18 \\
    & SCALECRAFTER    & 89.12 & 12.75 & 61.43 & 15.30 & 29.72 \\ 
    & MultiDiffusion  & 74.39 & 12.34 & 46.60 & 15.66 & \textbf{31.57} \\
    & DemoFusion      & 68.06 & 11.69 & \underline{46.32} & 16.66 & 29.75 \\
    & CutDiffusion      & 64.93    &15.74   &47.84   & 21.79  &   28.93\\ 

    & ASGDiffusion (Ours) & \textbf{64.27} & \textbf{15.98} & 46.95 & \textbf{22.51} & \underline{30.31} \\ \hline

    \multirow{6}{*}{2048$\times$2048} 
    & SDXL + BSRGAN   & 67.48 & \textbf{16.83} & \textbf{42.79} & 22.36 & \underline{30.64} \\ 
    & SCALECRAFTER    & 81.32 & 15.80 & 64.32 & 18.97 & 29.21 \\ 
    & MultiDiffusion  & 78.33 & 13.56 & 69.80 & 19.85 & 29.64 \\
    & DemoFusion      & \textbf{66.85} & \underline{16.59} & \underline{43.85} & \textbf{23.46} & 30.48 \\ 
    & CutDiffusion      & 71.04     & 15.30  &  45.47 & 21.19  &30.34   \\ 

    & ASGDiffusion (Ours) & \underline{68.49} & 16.23 & 46.10 & \underline{22.82} & \textbf{30.94} \\ \hline

    \multirow{6}{*}{3072$\times$3072} 
    & SDXL + BSRGAN   & \underline{69.35} & \underline{16.71} & \textbf{48.38} & \underline{19.01} & \underline{30.24} \\ 
    & SCALECRAFTER    & 89.16 & 12.46 & 87.95 & 13.03 & 28.11 \\ 
    & MultiDiffusion  & 101.44 & 9.62 & 74.61 & 15.42 & 29.74 \\
    & DemoFusion      & \textbf{64.85} & \textbf{17.11} & \underline{53.42} & \textbf{21.82} & \textbf{30.73} \\ 
    & CutDiffusion      & 71.97     & 12.49  & 63.43  &16.81   & 28.17  \\ 

    & ASGDiffusion (Ours) & 73.32 & 12.68 & 59.82 & 16.99 & 28.53 \\ \hline

    \end{tabular}
\end{table}

\subsection{Quantitative evaluation}
We quantitatively assess the model using the Laion-5B dataset~\cite{schuhmann2021laion}, utilizing 1,000 sampled captions for high-resolution image generation and a set of 10,000 real images. We evaluate image quality and semantic similarity using FID~\cite{heusel2017fid}, IS~\cite{salimans2016improved}, and CLIP Score~\cite{radford2021learning}. To evaluate high-resolution images more effectively, we compute FID$_c$ and IS$_c$ by cropping and resizing patches to a resolution of 1K, as suggested by~\cite{park2021benchmark}. Results are reported at three resolutions.

Tab.~\ref{tab:quantitative_comparison} presents the quantitative comparison of ASGDiffusion with other methods. At lower resolutions such as 1024 × 2048, ASGDiffusion demonstrates the best FID and IS scores, indicating superior image quality and diversity. At higher resolutions, particularly 3072 × 3072, DemoFusion outperforms ASGDiffusion in both FID and IS metrics. This discrepancy can be attributed to two main factors. Firstly, ASGDiffusion synthesizes high-resolution images by combining patches derived from the default resolution of the Latent Diffusion Model (LDM). As the target resolution increases, the number of necessary patches grows, resulting in less effective pixel interaction and a decline in global consistency among patches. Secondly, while DemoFusion uses a progressive upsampling strategy that more effectively preserves high-resolution details, ASGDiffusion directly upsamples from the original resolution to the target resolution. This direct upsampling method may lead to a loss of fine details, which further contributes to the performance gap at higher resolutions.

\subsection{Ablation studies}
\begin{figure}[t]
    \centering
    \includegraphics[width=\linewidth]{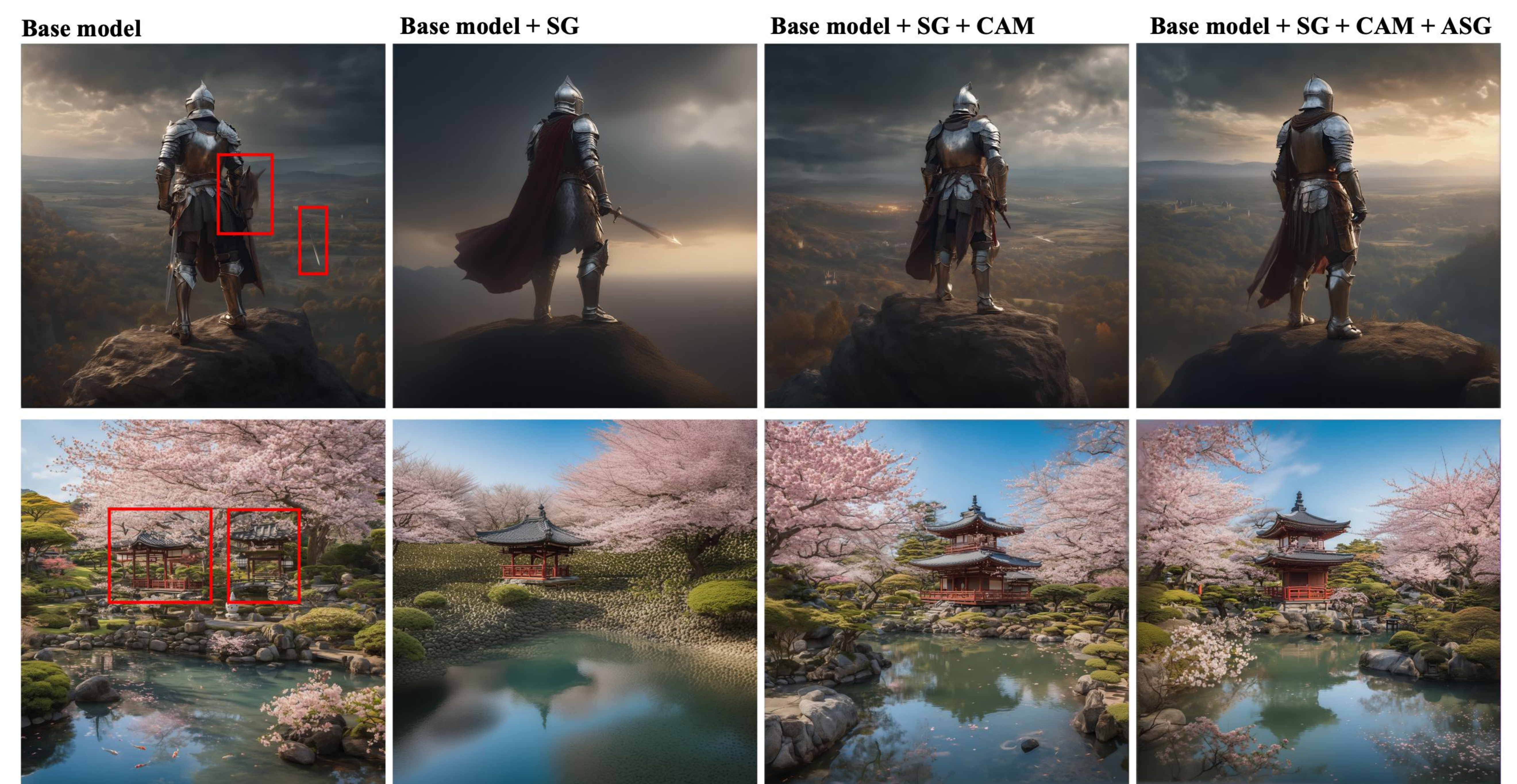}
    \caption{The ablation study of three components introduced in ASGDiffusion: SG (Structure Guidance), CAM (Cross-Attention Mask), and Asynchronous Structure Guidance (ASG). All results are presented at a resolution of 2048 × 2048.}
    \label{fig6}
\end{figure}

The aim of our ablation studies is to evaluate the effect of each key module in ASGDiffusion on the overall image quality. Specifically, we evaluate the contribution of three main components: Structure Guidance (SG), Cross-Attention Mask (CAM), and Asynchronous Structure Guidance (ASG). By adding each module progressively, we illustrate their individual and collective impact on the overall image generation process, as shown in Fig.~\ref{fig6}.

In the base model, significant issues were observed without guidance mechanisms, such as pattern repetition and inconsistent semantic structures. These artifacts occur due to a lack of effective structural guidance, resulting in semantic confusion and repetition. To address these issues, we introduced Structure Guidance (SG) to guide the generation of additional patches, ensuring a coherent global structure. This addition significantly enhances semantic consistency by eliminating repetitive patterns. However, we noticed that it resulted in background blurriness, particularly in areas like the sky or distant regions. To mitigate this, we added the Cross-Attention Mask (CAM). CAM utilizes the cross-attention heat map to create a mask that modulates the influence of structural guidance. It ensures that areas of high attention receive more guidance, while background regions are less influenced. Incorporating CAM greatly enhanced the clarity of the background, leading to well-balanced images that exhibit both consistent global structures and refined details. Finally, we introduced Asynchronous Structure Guidance (ASG) to improve efficiency. ASG uses guidance from the previous step ($t-1$) for the current step ($t$), experiments showed minimal quality differences between synchronous and asynchronous approaches, confirming ASG's effectiveness in maintaining image quality while reducing communication costs. This enables efficient parallel processing across multiple GPUs, accelerating the denoising process.

In summary, each module has a unique role: SG ensures global semantic consistency, CAM preserves background clarity, and ASG minimizes synchronization costs, thereby enhancing overall efficiency. Together, The components of ASGDiffusion work to efficiently generate high-quality, high-resolution images.

\begin{figure*}[ht]
    \centering
    \includegraphics[width=0.9\linewidth]{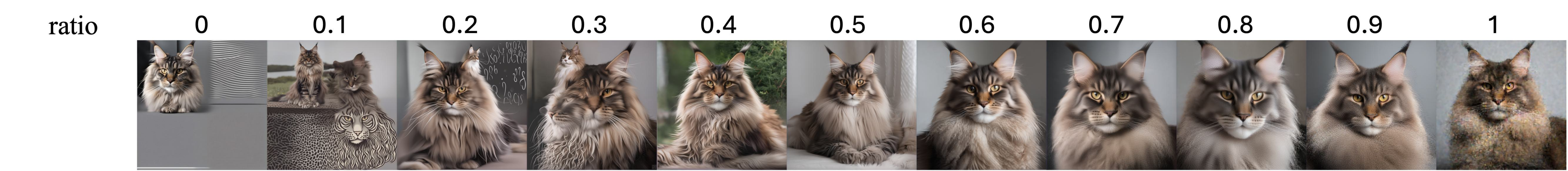}
    \caption{{The Effect of the Ratio of Global Semantic Structuring to Details Denoising on Image Quality.} 
   }
    \label{fig:ratio}


     \includegraphics[width=0.9\linewidth]{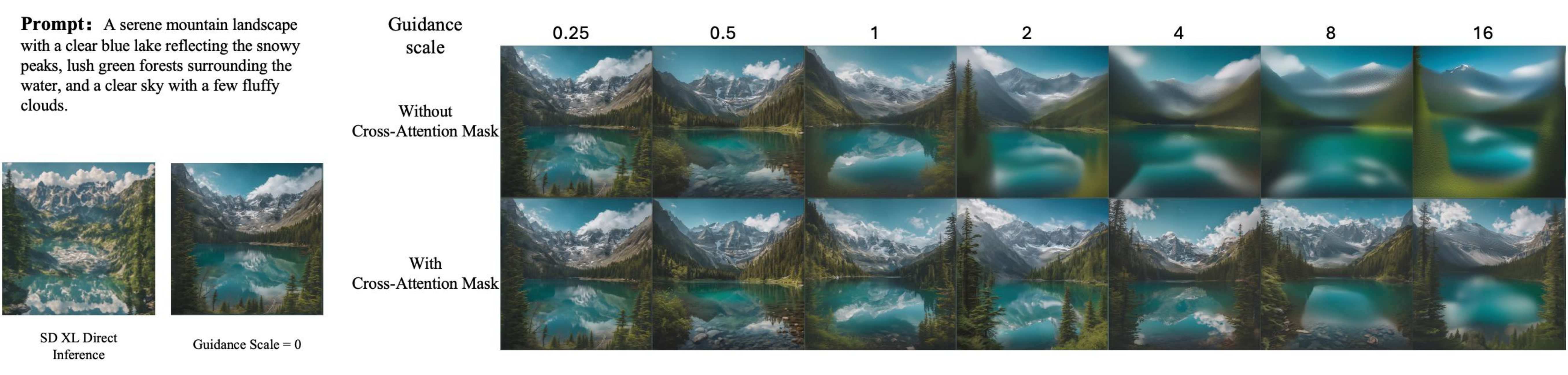}
    \caption{Analysis of the Impact of the Structure Guidance Scale on Image Generation.}
    \label{fig7}
\end{figure*}

\subsection{Analysis on hyperparameters}
\textbf{Key Patch Guidance Scale.} We conducted tests on image generation both with and without the attention mask, using various guidance values. As illustrated in Fig.~\ref{fig7}, the images produced without the attention mask became progressively blurrier as the guidance strength increased. Conversely, introducing the attention mask greatly reduced blurriness and enhanced overall image quality.

\textbf{Ratio of Global Semantic Structuring to Details Denoising Steps.}
We tested the impact of the ratio between the two stages on image quality. The experimental results are shown in Fig.~\ref{fig:ratio}. We found that a very low ratio in the first stage resulted in chaotic outputs. As the ratio increased, the image quality improved; however, an excessively high ratio introduced checkerboard artifacts while still preserving the semantic structure.

\section{Discussion}

\textbf{Human Evaluation.} Relying only on quantitative metrics fails to fully capture the quality of images produced by the model. To evaluate the image quality, we conducted experiments designed to assess it subjectively. We presented images generated by ScaleCrafter, DemoFusion, and ASGDiffusion in a randomized order, all created using the same prompts and resolutions. Participants were asked to rank the images according to their personal perceptions. To conduct this experiment, we recruited 20 volunteers. Our statistical results, as shown in Tab.~\ref{tab:user_study_ranking}, reveal that images generated by ASGDiffusion significantly outperform those produced by ScaleCrafter and DemoFusion, highlighting the exceptional performance of our method. Additional details and analyses are provided in the supplementary material.

\begin{table}[htbp]
\vspace{-0.15cm}
\small
    \centering
    \caption{The results of the average ranking human evaluation were based on metrics of visual appeal and text fidelity, assessed by twenty volunteer participants. Lower ranking numbers signify better performance for the corresponding method.}
    \vspace{-0.25cm}
    \label{tab:user_study_ranking}
    \begin{tabular}{lccc}
    \hline
    \scriptsize Method & \scriptsize SCALECRAFTER & \scriptsize DemoFusion  & \scriptsize ASGDiffusion (Ours)\\ \hline
    \scriptsize Rank$\downarrow$   & \scriptsize 2.11         & \scriptsize 1.97         & \textbf{\scriptsize 1.68 }           \\ \hline
    \end{tabular}
    \vspace{-0.25cm}
\end{table}

\begin{figure}
    \centering
    \includegraphics[width=\linewidth]{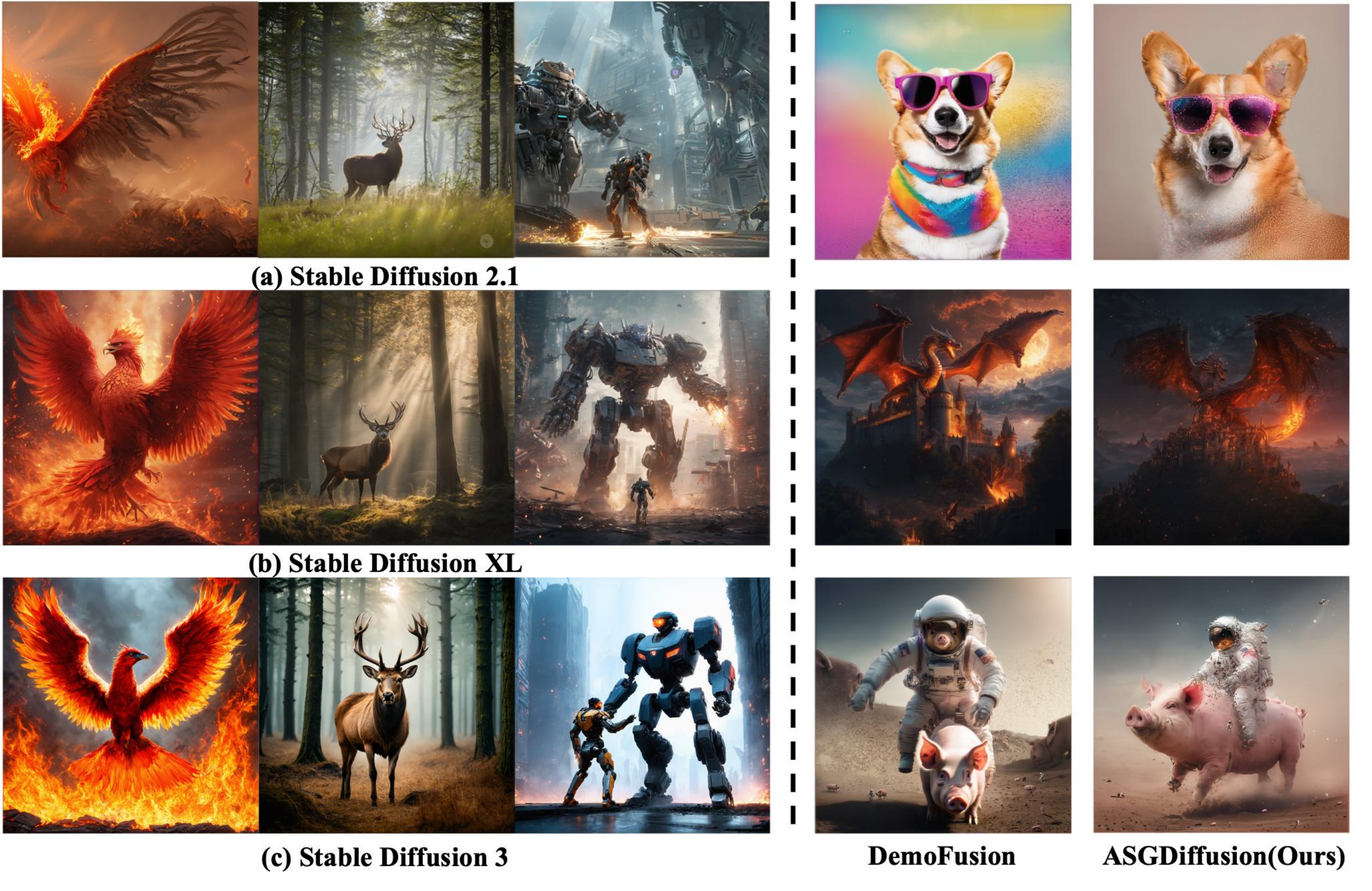}
    \vspace{-0.4cm}
    \caption{Left: Experimental results of our method on other versions of diffusion models, 4x upsample; Right: Failure cases, 16x upsample.}
    \vspace{-0.6cm}
    \label{fig10}
\end{figure}

\textbf{Limitation.}
ASGDiffusion has several limitations that need to be addressed.
(1) Our method faces challenges with the repetition of small objects when generating ultra-high-resolution images (4096x4096), as illustrated in Fig.~\ref{fig10}. ASGDiffusion synthesizes images by combining patches from the default resolution of the Latent Diffusion Model (LDM). As the resolution increases, more patches are needed, which leads to less effective pixel interaction and decreased consistency across patches. In the future, using a progressive upsampling method could help address this limitation.
(2) While we used cross-attention mask to minimize image blur, as shown in Fig.~\ref{fig10}, the dog's head was generated clearly, but some blur remained on its body. We believe this is because the attention mask does not fully cover the dog's body. Future research could tackle this issue by utilizing a more accurate mask.
(3) Since ASGDiffusion is a training-free high-resolution image generation model, its performance is inherently limited by the capabilities of the underlying LDM. We tested our method on different versions of diffusion models, as illustrated in Fig.~\ref{fig10}. Applying our method to more generation models is also promising in the future.

\section{Conclusion}
ASGDiffusion is a method for generating high-resolution images without training, addressing issues of pattern repetition, and reducing computational costs. Using structure guidance with a cross-attention mask ensures semantic consistency while reducing repetitive artifacts. We also propose a parallelism strategy to make the structure guidance asynchronous, which minimizes generation time and memory usage. ASGDiffusion delivers outstanding results in generating high-resolution images both effectively and efficiently.

{
    \small
    \bibliographystyle{ieeenat_fullname}
    \bibliography{main}
}

\clearpage
\setcounter{page}{1}
\maketitlesupplementary

\section*{Overview}

The following aspects are included in this supplementary material:

\begin{itemize}
    \item \textbf{Supplementary experimental analysis}
    \begin{itemize}
        \item Hyperparameter experiments:
        
        \item User study results:
       
    \end{itemize}

    \item \textbf{Prompts used for image generation}
    \begin{itemize}
        \item List of prompts associated with each figure in the main text.
    \end{itemize}

    \item \textbf{Additional visualization results}
    \begin{itemize}
        \item High-resolution outputs generated by ASGDiffusion.
    \end{itemize}

\end{itemize}
\section*{Detailed Hyperparameter Experiments}

\subsection*{Experimental Setup}

In this section, we explore the impact of two critical hyperparameters: the ratio of denoising steps allocated to Global Semantic Structuring (denoted as T1 / (T1 + T2)) and the guidance scale used in Structure Guidance. The experiments are conducted using a broad range of values for these hyperparameters to comprehensively analyze their effects.We present the detailed experimental results in the Figure~\ref{fig1}.

\begin{itemize}
    \item \textbf{Ratio of Global Semantic Structuring to Detail Denoising (T1 / (T1 + T2))}: 
    We tested the following values for the ratio: 0.0, 0.1, 0.2, 0.3, 0.4, 0.5, 0.6, 0.7, 0.8, 0.9, 1.0. These values cover a spectrum from full focus on Detail Denoising (0.0) to full focus on Global Semantic Structuring (1.0).

    \item \textbf{Guidance Scale for Structure Guidance}:
    The guidance scale values tested are: 0, 0.5, 1, 2, 4, 8, 16. This scale controls the strength of the guidance provided by the key patch during denoising.

    \item \textbf{Model Version}:
    All experiments were conducted using the Stable Diffusion XL (SD XL) model.

    \item \textbf{Resolution}:
    Images were generated at a resolution of 2048x2048 pixels.

    \item \textbf{Hardware}:
    The experiments were performed on an NVIDIA RTX 4090 GPU.
\end{itemize}

\subsection*{Results Comparison}

Through the generated images illustrated in Figure~\ref{fig1}, we observe the following trends:

\begin{itemize}
    \item \textbf{Impact of T1 / (T1 + T2) Ratios}: Low ratios (e.g., 0.0 or 0.1) result in chaotic global structures due to insufficient Global Semantic Structuring. High ratios (e.g., 0.9 or 1.0) improve global coherence but can introduce checkerboard artifacts and reduce detail quality.

    \item \textbf{Impact of Guidance Scales}: Low guidance scales (e.g., 0.0) lead to inconsistent Detail Denoising, while moderate scales (e.g., 2.0 or 4.0) offer a good balance between structure and detail. Very high scales (e.g., 16.0) result in overly smooth and artificial images.

    \item \textbf{Extreme Settings}: Combining high ratios with high guidance scales can cause images to lose natural texture, while low ratios with low guidance scales produce incoherent and disorganized images.
\end{itemize}

\subsection*{Discussion and Conclusion}

The choice of hyperparameters significantly impacts the quality of high-resolution image generation. For most use cases, a balanced approach—where the ratio T1 / (T1 + T2) is around 0.5 and the guidance scale is moderate (e.g., 1.0 or 2.0)—provides the best trade-off between global structure coherence and local Detail Denoising.

\textbf{Future Work}:
Further testing on different resolutions and image types is needed to optimize these hyperparameters across a wider range of applications. Exploring adaptive methods for dynamically adjusting these settings during the generation process could also enhance performance.

This analysis offers insights into the delicate balance between structure and detail, guiding future research and practical applications in high-resolution image generation.

\section*{User Study}

The user study involved 20 participants who were asked to evaluate 50 images per method. Participants used a custom evaluation interface, as described in the main paper. They ranked images generated by three different methods: DemoFusion, ASGDiffusion, and ScaleCrafter, focusing on visual appeal and fidelity to the prompt.

The results are summarized as follows:

\begin{figure*}[t]
    \centering
    
    \includegraphics[width=0.8\textwidth]  {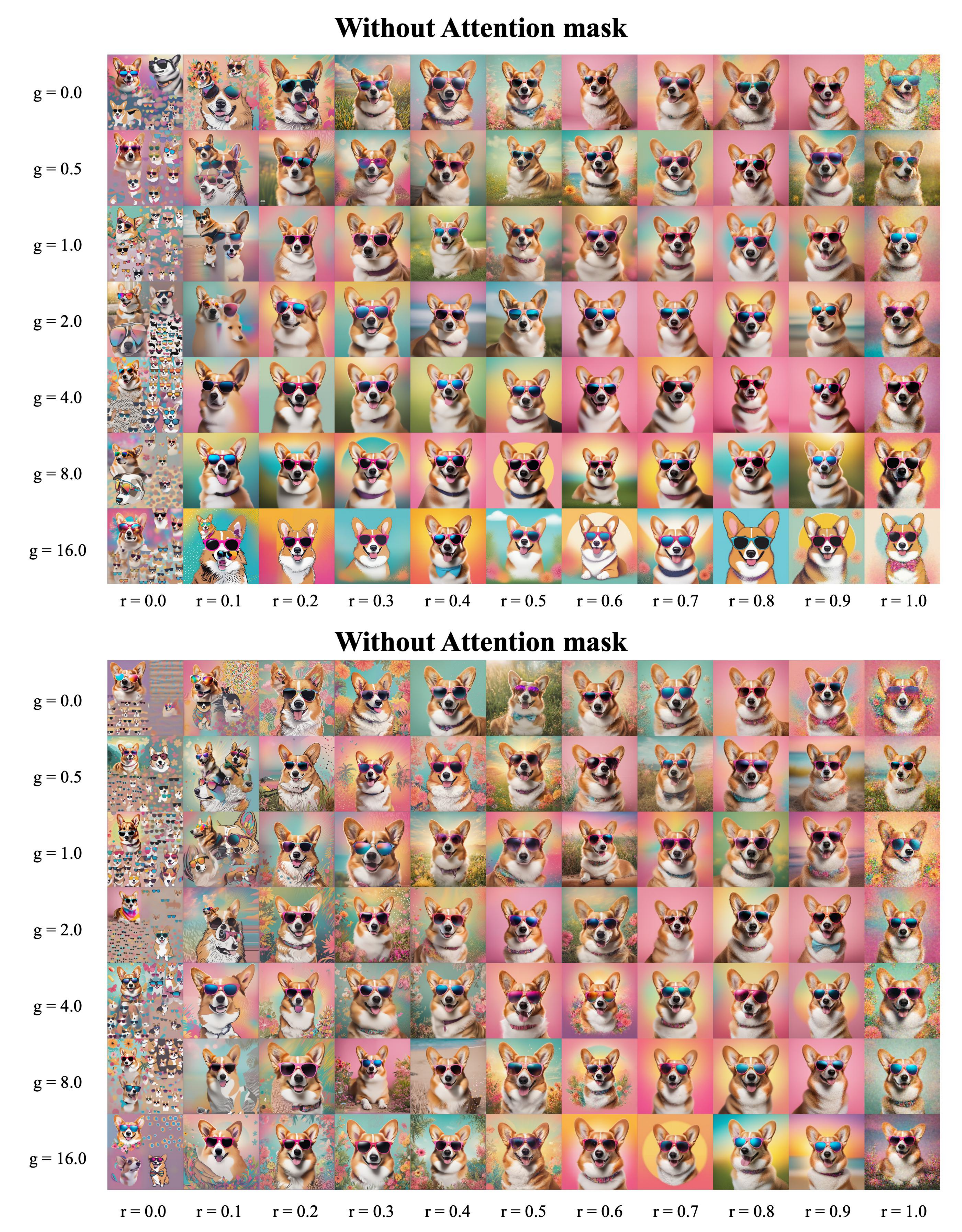} 
    \caption{The figure illustrates the effects of varying guidance scale (g) and ratio (r) settings on the generated images. The top grid shows the results without using an attention mask, while the bottom grid shows the results with the attention mask applied. Each cell in the grids corresponds to a specific combination of guidance scale (g) and ratio (r), where g varies from 0.0 to 16.0 (vertical axis) and r varies from 0.0 to 1.0 (horizontal axis).}
    \label{fig1}
    
\end{figure*}

\begin{itemize}
\item \textbf{Overall Preference Scores}: The mean preference scores for DemoFusion, ASGDiffusion, and ScaleCrafter were 1.969, 1.683, and 2.206, respectively. Lower scores indicate higher preference.
\begin{table}[htbp]
\vspace{-0.15cm}
\small
    \centering
    \caption{The average ranking human evaluation results based on visual appeal and text fidelity metrics assessed by twenty volunteer participants. Lower ranking numbers indicate better performance of the corresponding method.}
    \vspace{-0.25cm}
    \label{tab:user_study_ranking}
    \begin{tabular}{lccc}
    \hline
    \scriptsize Method & \scriptsize SCALECRAFTER & \scriptsize DemoFusion  & \scriptsize ASGDiffusion (ours)\\ \hline
    \scriptsize Rank$\downarrow$   & \scriptsize 2.11         & \scriptsize 1.97         & \textbf{\scriptsize 1.68 }           \\ \hline
    \end{tabular}
    \vspace{-0.25cm}
\end{table}
\item \textbf{Win Rates}: ASGDiffusion demonstrated a win rate of 60.10\% against DemoFusion and 67.51\% against ScaleCrafter, indicating a strong overall preference for ASGDiffusion.

\begin{figure}[ht]
    \centering
    \includegraphics[width=0.8\linewidth]{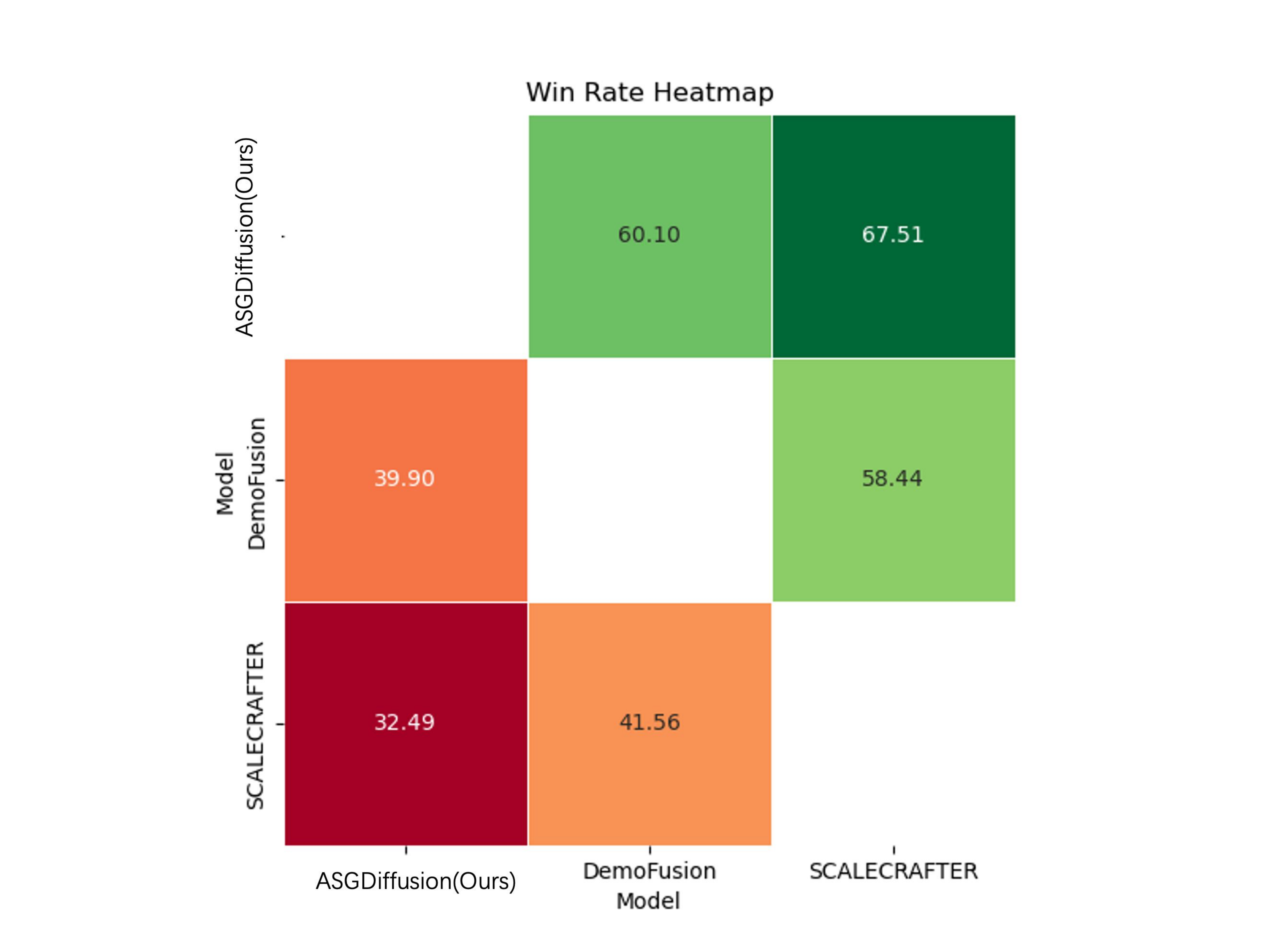}
    \caption{The heatmap shows the win rates of each method in the user study. Participants ranked images generated by different models based on visual appeal and fidelity to the prompt, with ASGDiffusion demonstrating the highest win rates against both DemoFusion and ScaleCrafter.}
    \label{fig:user_study_interface}
\end{figure}

\item \textbf{Head-to-Head Comparisons}: ASGDiffusion won in 470 instances against DemoFusion, while losing in 312 instances. Against ScaleCrafter, ASGDiffusion won in 528 instances and lost in 254 instances.
\begin{figure}[ht]
    \centering
    \includegraphics[width=0.8\linewidth]{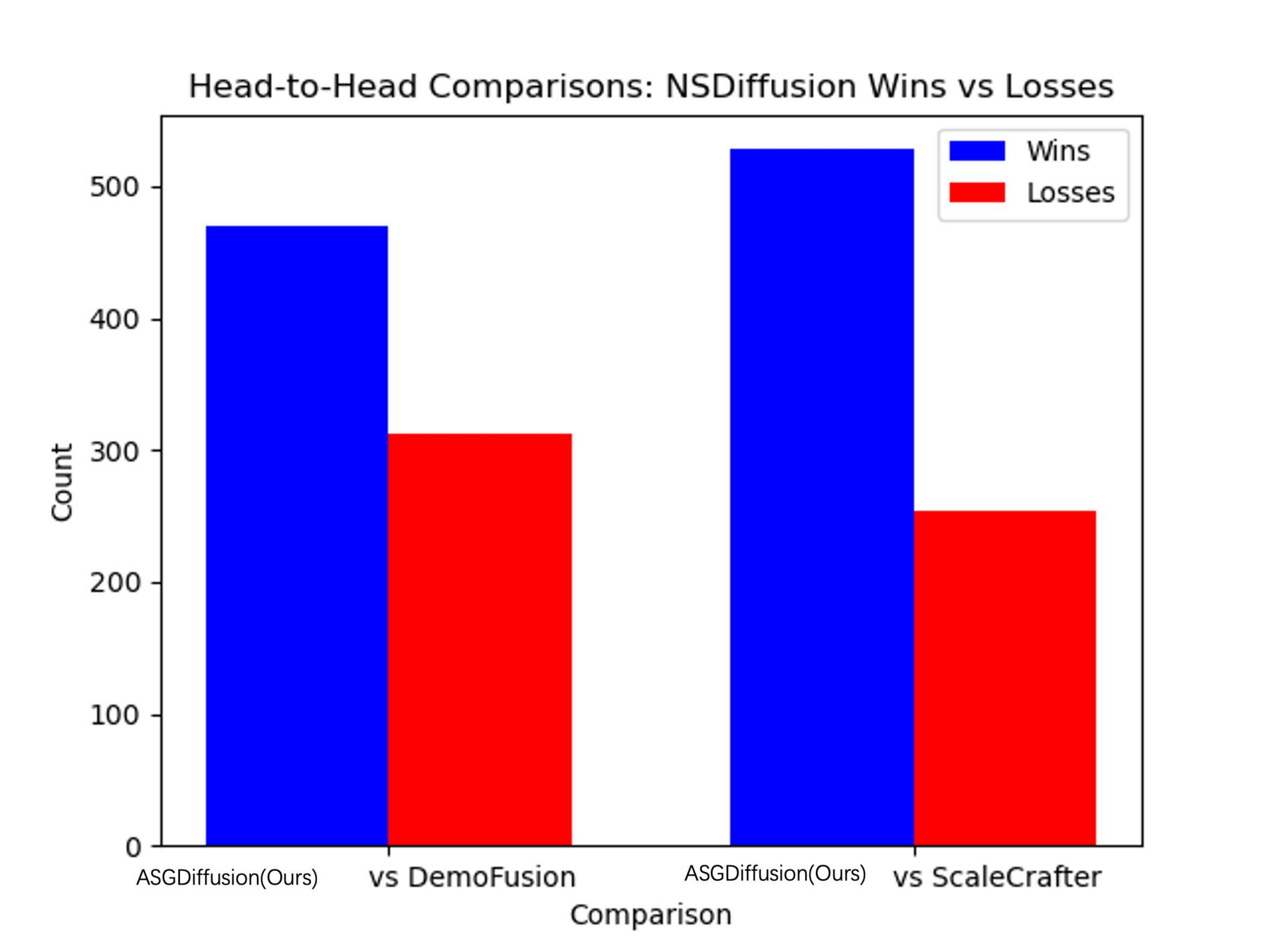}
    \caption{Head-to-Head comparison between ASGDiffusion and other methods showing the number of wins and losses in the user study.}
    \label{fig:head_to_head_comparison}
\end{figure}
\end{itemize}

These results indicate that ASGDiffusion was generally favored over the other methods, especially in its ability to generate visually appealing and faithful representations of the given prompts.

\subsection*{Discussion}

In summary, the human evaluation revealed that ASGDiffusion consistently outperformed both ScaleCrafter and DemoFusion in generating high-quality images. This was determined through a user study involving 20 volunteers who ranked images based on subjective perceptions. The results clearly indicate the superiority of ASGDiffusion, confirming the effectiveness of our method in producing visually appealing and faithful representations.

\section*{Prompts Used for Image Generation}   

\subsubsection*{Figure 1 in the main text}

\begin{itemize}
    \item A futuristic soldier in high-tech armor, standing in a war-torn city, in the style of sci-fi action art, gritty textures, dark atmosphere, ultra-detailed, photorealistic, 8k resolution, cinematic.
    \item Urban Jungle Shaman: A bold shaman in urban jungle attire, navigating a vibrant cityscape. His makeup features earthy tones with intricate tribal patterns and feather details. The city is filled with towering buildings, lush greenery, and vibrant street art. Liquid paint and urban foliage merge and interact, creating a dynamic and adventurous atmosphere.
    \item A mystical wizard casting a spell in an ancient library, in the style of fantasy illustration, detailed bookcases, magical energy swirling, dark atmosphere, 8k resolution, cinematic.
    \item A whimsical fairytale village, with candy-colored houses and cobblestone streets, in the style of children’s book illustrations, vibrant colors, playful details, magical atmosphere, 8k resolution, trending on artstation.
    \item A neon-lit cyberpunk street scene, with rain-soaked pavement reflecting colorful signs, in the style of dystopian sci-fi, gritty textures, dark atmosphere, photorealistic, 8k resolution, cinematic.
    \item A tranquil lakeside cabin at sunset, with mountains in the background, in the style of romantic landscape painting, soft golden light, detailed wood textures, peaceful atmosphere, ultra-high definition, photorealistic, 8k.
    \item A tranquil autumn forest with golden leaves falling, in the style of romantic landscape painting, soft golden light, detailed foliage, peaceful atmosphere, ultra-high definition, photorealistic, 8k. 
    \item A vintage 1950s diner at night, neon signs glowing, in the style of retro Americana, detailed textures, nostalgic atmosphere, photorealistic, ultra-high definition, 8k resolution.
    \item A steampunk airship sailing through the clouds, gears and cogs exposed, in the style of Victorian science fiction, rich metallic textures, detailed engineering, sunset sky, dramatic lighting, 8k resolution, cinematic.    
    \item A vibrant underwater scene with colorful coral reefs and exotic fish, in the style of marine life photography, detailed textures, bright colors, serene atmosphere, ultra-high definition, photorealistic, 8k.

\end{itemize}

\subsubsection*{Figure 2 in the main text}

\begin{itemize}
    \item A corgi wearing cool sunglasses, with a colorful summer background.
    
\end{itemize}

\subsubsection*{Figure 3 in the main text}

\begin{itemize}
    \item A squirrel eating an acorn.

\end{itemize}

\subsubsection*{Figure 4 in the main text}

\begin{itemize}
    \item A squirrel eating an acorn.
    
\end{itemize}

\subsubsection*{Figure 5 in the main text}

\begin{itemize}
    \item An astronaut riding a pig, highly realistic DSLR photo, cinematic shot.
    \item A fox peeking out from behind a bush, with a forest clearing in the background.
    \item A young badger with a rose.
    \item A corgi wearing cool sunglasses, with a colorful summer background.
    \item Astronaut on Mars During sunset.
    \item A tranquil orchard with fruit trees in bloom.
    
\end{itemize}

\subsubsection*{Figure 6 in the main text}

\begin{itemize}
        \item A squirrel eating an acorn.

\end{itemize}

\subsubsection*{Figure 7 in the main text}

\begin{itemize}
    \item A knight in shining armor, standing on a cliff overlooking a battlefield, in the style of Renaissance art, dramatic lighting, detailed armor, heroic pose, epic atmosphere, 8k resolution, oil painting texture.
    \item A serene temple surrounded by cherry blossoms.

\end{itemize}

\subsubsection*{Figure 8 in the main text} 

\begin{itemize}
    \item A fluffy Maine Coon cat

\end{itemize}

\subsubsection*{Figure 9 in the main text}

\begin{itemize}
    \item A serene mountain landscape with a clear blue lake reflecting the snowy peaks, lush green forests surrounding the water, and a clear sky with a few fluffy clouds.

\end{itemize}

\subsubsection*{Figure 10 in the main text}

\begin{itemize}
    \item A mythical phoenix rising from the ashes, with flames swirling around it, in the style of classical mythology, vibrant red and orange colors, detailed feathers, dynamic pose, epic atmosphere, 8k resolution.
    \item A majestic stag standing in a misty forest, with sunlight filtering through the trees, in the style of wildlife photography, detailed fur textures, ethereal atmosphere, photorealistic, 8k resolution, cinematic.
    \item A cybernetic warrior battling a giant robot in a futuristic city, in the style of sci-fi action art, intense action, detailed character design, dynamic composition, 8k resolution, cinematic.
    \item A corgi wearing cool sunglasses, with a colorful summer background.
    \item A majestic dragon soaring over a medieval castle, with fiery breath lighting up the night sky, in the style of classic fantasy art, ultra-detailed scales, vibrant flames, moonlit scene, 8k resolution, trending on artstation.
    \item An astronaut riding a pig, highly realistic DSLR photo, cinematic shot.

\end{itemize}

\section*{Additional Generated Images}
\begin{figure*}[t]
    \centering
    \includegraphics[width=\linewidth]{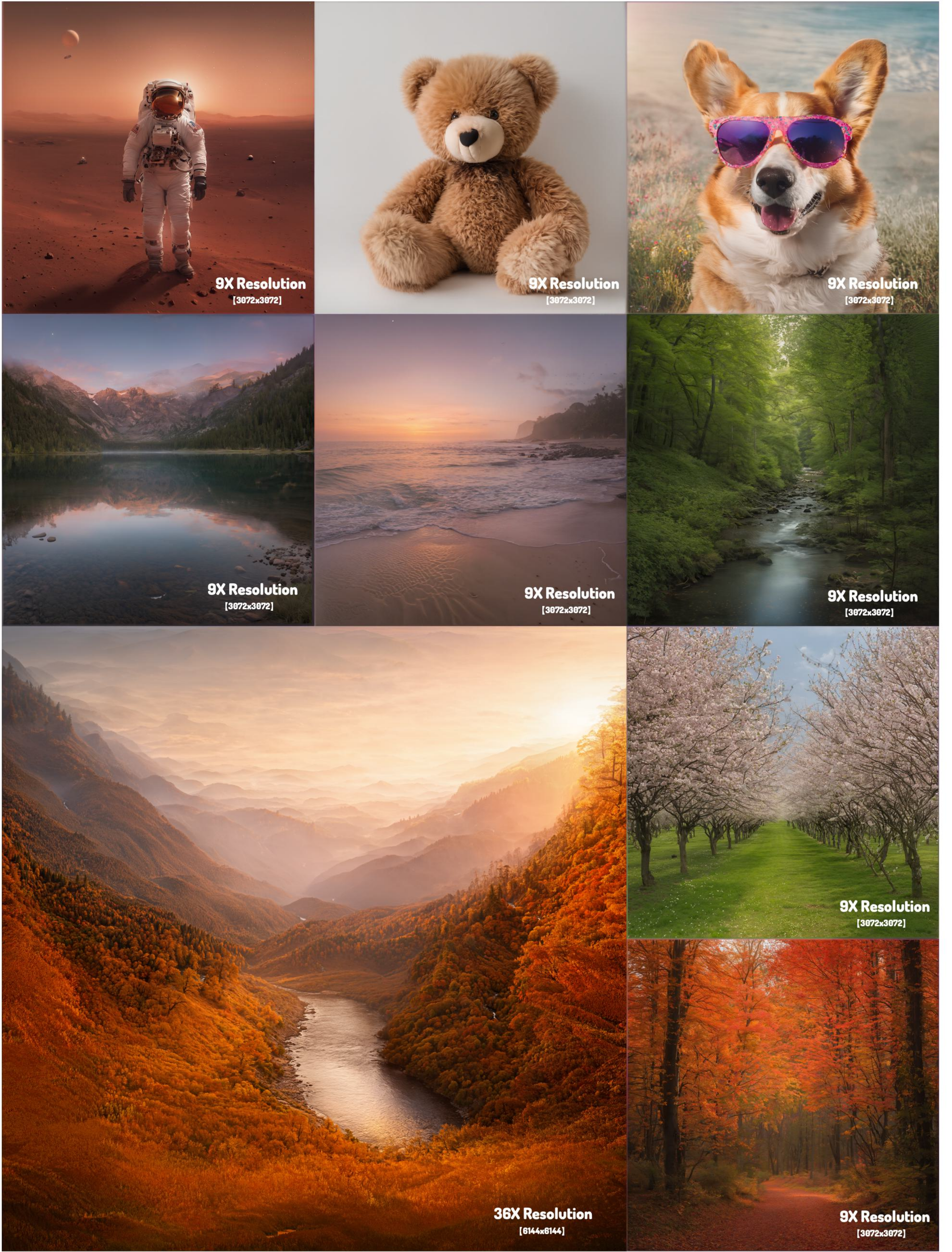}
    \caption{Additional generated results using ASGDiffusion.}
    \label{addition1}

\end{figure*}

\begin{figure*}[t]
    \centering
    \includegraphics[width=0.9\linewidth]{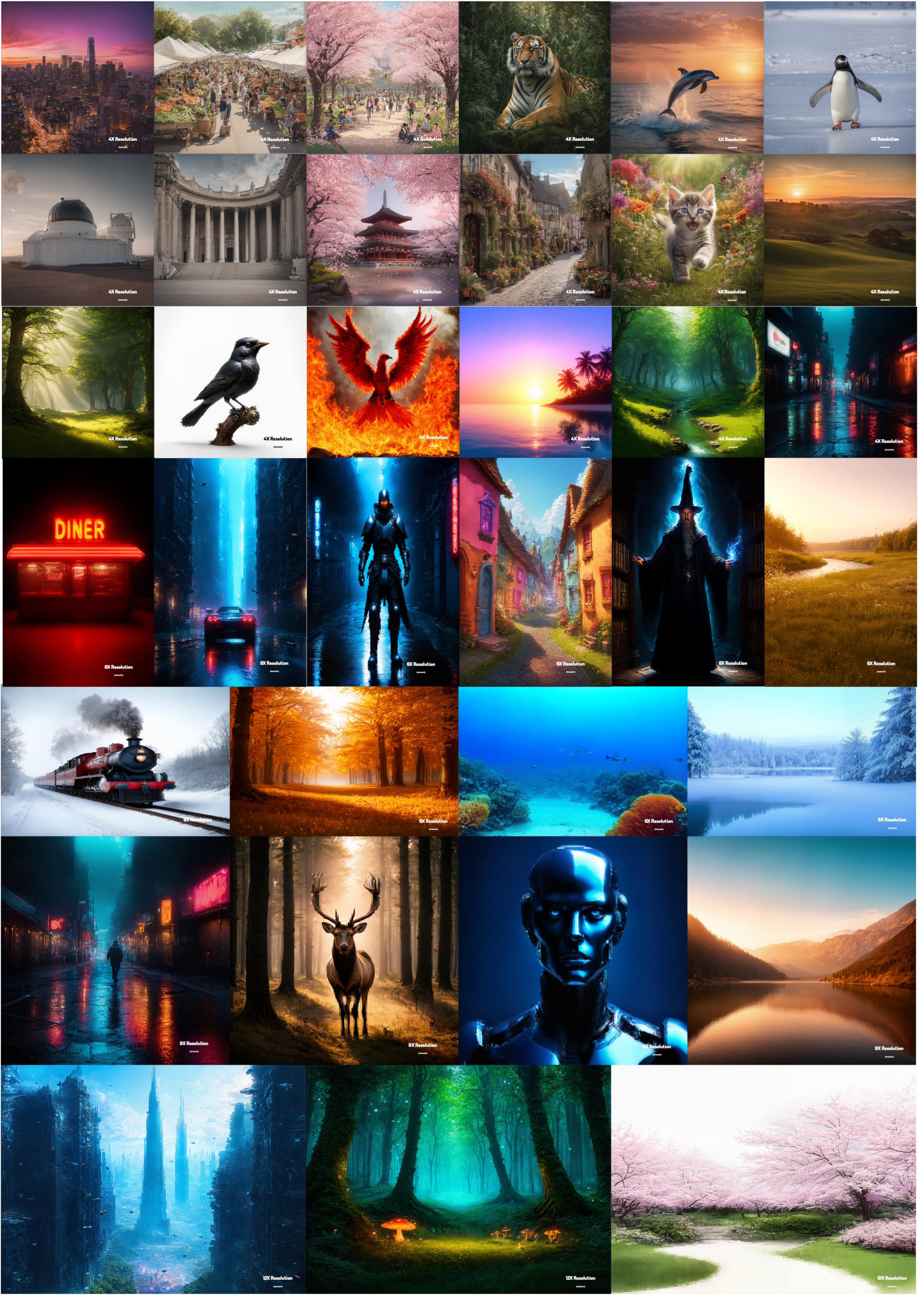}
    \caption{Additional generated results using ASGDiffusion.}
    \label{addition2}
\end{figure*}

In this section, we present additional generated images using ASGDiffusion. Figure~\ref{addition1} and~\ref{addition2} shows a variety of outputs under different resolutions, further demonstrating the model's robustness and ability to produce high-quality images across various scenarios. These results highlight the diversity and consistency of ASGDiffusion in generating visually appealing and faithful representations.

\end{document}